%% file: main.tex
\documentclass[10pt,twocolumn,letterpaper]{article}

\usepackage[pagenumbers]{iccv} 

\input{preamble}

\definecolor{iccvblue}{rgb}{0.21,0.49,0.74}
\usepackage[pagebackref,breaklinks,colorlinks,allcolors=iccvblue]{hyperref}

\def\paperID{*****} 
\def\confName{ICCV}
\def\confYear{2025}

\title{Generative AI for Cel-Animation: A Survey}

\author{%
Yunlong Tang\textsuperscript{1},
Junjia Guo\textsuperscript{1},
Pinxin Liu\textsuperscript{1},
Zhiyuan Wang\textsuperscript{2},
Hang Hua\textsuperscript{1},
Jia-Xing Zhong\textsuperscript{3},\\
Yunzhong Xiao\textsuperscript{4},
Chao Huang\textsuperscript{1},
Luchuan Song\textsuperscript{1},
Susan Liang\textsuperscript{1},
Yizhi Song\textsuperscript{5},
Liu He\textsuperscript{5},\\
Jing Bi\textsuperscript{1},
Mingqian Feng\textsuperscript{1},
Xinyang Li\textsuperscript{1},
Zeliang Zhang\textsuperscript{1},
Chenliang Xu\textsuperscript{1}\\[2pt]
\textsuperscript{1}University of Rochester ,
\textsuperscript{2}UCSB ,
\textsuperscript{3}University of Oxford ,
\textsuperscript{4}CMU ,
\textsuperscript{5}Purdue University\\[2pt]
{\tt\small\url{https://github.com/yunlong10/Awesome-AI4Animation}}
}

\begin{document}
\maketitle
\begin{abstract}
\input{sec/0_abs}
\end{abstract}

\input{sec/1_intro}

\input{sec/2_pre}
\input{sec/3_methods}
\input{sec/4_case_study}

\input{sec/5_discussion}
\input{sec/6_conclusion}
{
    \small
    \bibliographystyle{ieeenat_fullname}
    \bibliography{main}
}


\end{document}

%% file: preamble.tex
\usepackage{amsmath,amsfonts}
\usepackage{algorithmic}
\usepackage{algorithm}
\usepackage{array}
\usepackage[font=normalsize,labelfont=sf,textfont=sf]{subcaption}
\usepackage{textcomp}
\usepackage{stfloats}
\usepackage{xcolor}
\usepackage{adjustbox}
\usepackage{tikz}
\usepackage{url}
\usepackage{bm}
\usepackage[edges]{forest}
\usepackage{booktabs}
\usepackage{multirow}
\usepackage{multicol}
\usepackage{makecell}
\usepackage{verbatim}
\usepackage{graphicx}
\usepackage{colortbl} 
\usepackage{pifont}
\usepackage{fontawesome5}
\usepackage{enumitem}
\usepackage{longtable}
\usepackage{tcolorbox}

\definecolor{mattered}{RGB}{214, 26, 60}
\definecolor{mattegreen}{HTML}{369F39}

\usepackage{bbding}

\definecolor{hidden-draw}{RGB}{20,68,106}
\definecolor{hidden-pink}{RGB}{255,245,247}
\definecolor{my_green}{RGB}{51,102,0}
\definecolor{my_red}{RGB}{204, 0, 0}
\definecolor{blue_color}{RGB}{0, 76, 153}   
\definecolor{red_color}{RGB}{179, 25, 25}   

\DeclareMathOperator*{\argmax}{arg\,max}

\definecolor{preproduction}{RGB}{46, 151, 78}
\definecolor{production}{RGB}{239, 120, 24}
\definecolor{postproduction}{RGB}{114, 97, 171}
\newcommand{\insightbox}[1]{%
    \begin{tcolorbox}[colframe=black!70, colback=blue!5, boxrule=1pt, arc=2mm]
        \includegraphics[width=0.35cm]{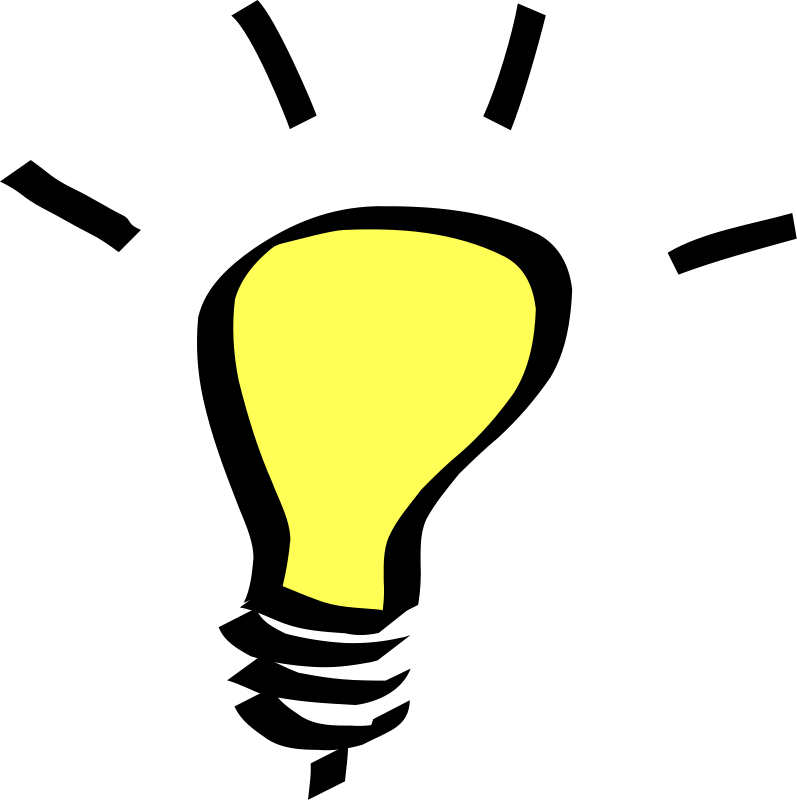}
        \textbf{\small#1}
    \end{tcolorbox}
}

%% file: sec/0_abs.tex
Traditional Celluloid (Cel) Animation production pipeline encompasses multiple essential steps, including storyboarding, layout design, keyframe animation, inbetweening, and colorization, which demand substantial manual effort, technical expertise, and significant time investment. 
These challenges have historically impeded the efficiency and scalability of Cel-Animation production. 
The rise of generative artificial intelligence (GenAI), encompassing large language models, multimodal models, and diffusion models, offers innovative solutions by automating tasks such as inbetween frame generation, colorization, and storyboard creation. 
This survey explores how GenAI integration is revolutionizing traditional animation workflows by lowering technical barriers, broadening accessibility for a wider range of creators through tools like AniDoc, ToonCrafter, and AniSora, and enabling artists to focus more on creative expression and artistic innovation. 
Despite its potential, challenges like visual consistency, stylistic coherence, and ethical considerations persist. Additionally, this paper explores future directions and advancements in AI-assisted animation. 


%% file: sec/1_intro.tex
\section{Introduction}

\begin{figure}[!ht]
    \centering
    \includegraphics[width=1\linewidth]{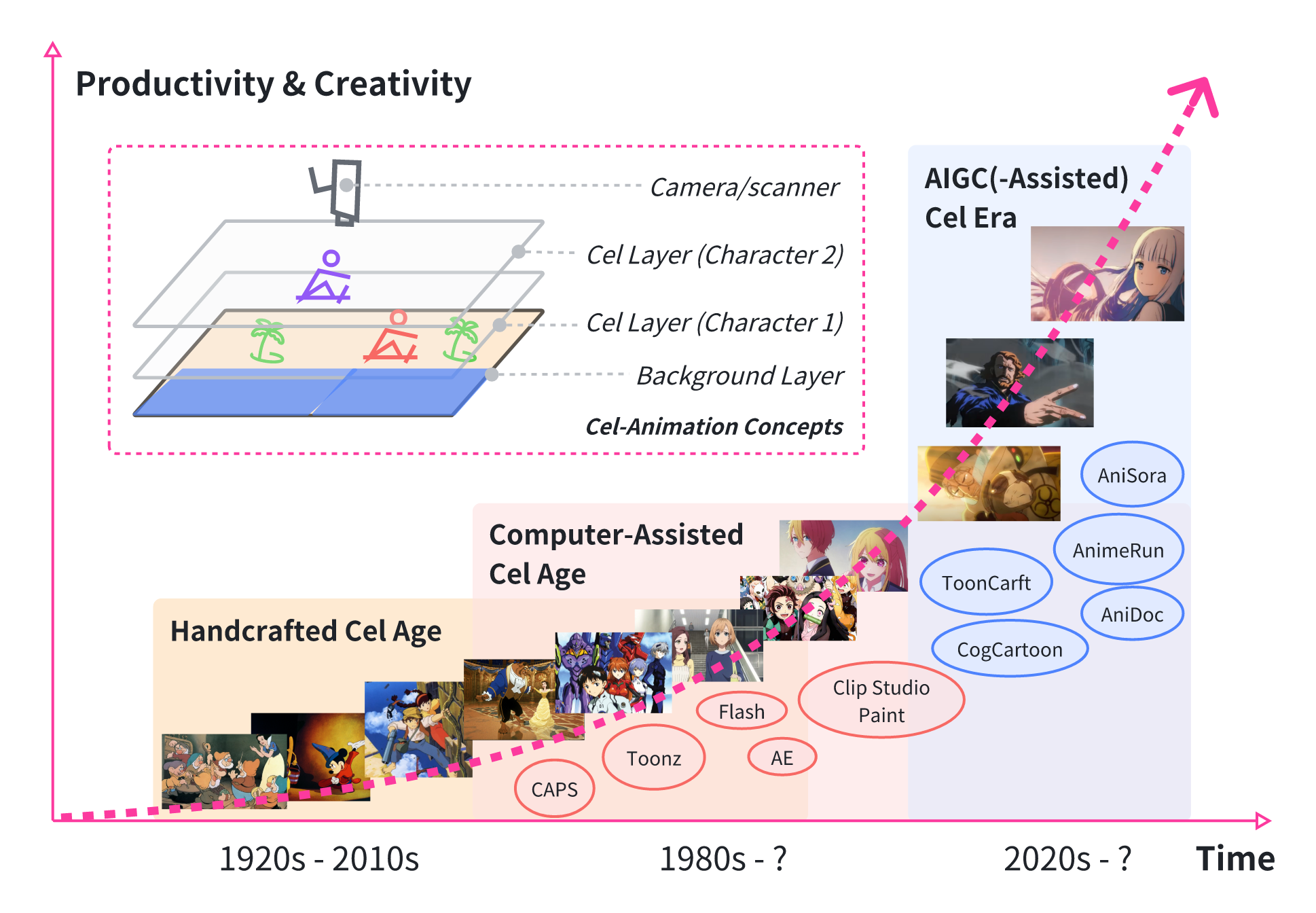}
    \caption{The three major phases in Cel-Animation history: the Handcrafted Cel Age (1920s-2010s), the Computer-Assisted Cel Age (1980s-present), and the emerging AIGC Cel Era (2020s onward). A layered structure of Cel-Animation is also shown.}
    \label{fig:teaser}
    \vspace{-1.5em}
\end{figure}

Animation, as a powerful medium for storytelling and artistic expression, has evolved significantly over the past century.
Celluloid (Cel) animation, established as a cornerstone of traditional animation in the early 1920s, has shaped the foundation of the modern animation industry through its distinctive frame-by-frame approach that combines artistic vision with technical precision.
This methodology laid the foundation for modern animation while highlighting a fundamental tension: the challenges of maintaining high artistic quality while improving production efficiency:
1) \textit{Time-intensive manual labor}: The creation of keyframes, inbetweening, and colorization often requires extensive manual effort. 
2) \textit{Technical complexity}: Coordinating multiple stages, from scripts to storyboard, from keyframe animation to coloring, involves a high degree of expertise and meticulous planning.
3) \textit{Creativity constraints}: Repetitive and labor-intensive tasks can detract from the time and energy animators devote to higher-level creative decisions.
These challenges pose barriers to efficiency and scalability, particularly in modern animation workflows where production timelines are tightening.

Recent breakthroughs in generative artificial intelligence (GenAI), including large language models (LLMs)~\cite{touvron2023llama,yang2024qwen2,le2023bloom}, multimodal LLMs (MLLMs)~\cite{liu2024visual, achiam2023gpt, tang2023video,Qwen2VL,chen2024far,zhu2023minigpt,hua2024mmcomposition,hua2024finecaption,yu2024promptfix}, and diffusion models~\cite{ho2020ddpm, song2020denoising, zhang2023controlnet, esser2024scaling, blattmann2023align}, offer promising solutions to these challenges. By automating repetitive and technically demanding tasks, GenAI frees animators to engage more deeply in high-level creative processes. To contextualize how GenAI reshapes animation workflows, we review Cel-Animation’s evolution through three historical phases, as illustrated in~\Cref{fig:teaser}:

\paragraph{The Handcrafted Cel Age (1920s-2010s)}
The introduction of celluloid sheets enabled animators to separate moving characters from static backgrounds, significantly enhancing scene complexity and allowing efficient reuse of backgrounds~\cite{studiobinder_cel_animation}. Core animation techniques developed during this era include storyboarding, keyframe drawing, manual inbetweening, and hand-painting of individual cels. Landmark works, such as Disney’s \textit{Snow White and the Seven Dwarfs} (1937), exemplify both the artistic potential and the inherent limitations of this approach: the film required over 200K hand-drawn frames and extensive human labor, highlighting issues of scalability and consistency.

\paragraph{The Computer-Assisted Cel Age (1980s-present).}
Advancements in digital technology significantly altered traditional Cel-Animation pipelines from the 1980s onward. Notably, Disney’s Computer Animation Production System (CAPS)~\cite{caps_wikipedia}, developed jointly with Pixar, automated coloring and compositing, achieving higher precision and complexity exemplified by films such as \textit{Beauty and the Beast} (1991) and \textit{The Lion King} (1994). Concurrently, Japanese studios leveraged tools such as Toonz~\cite{toonz_wikipedia}, OpenToonz~\cite{opentoonz}, Adobe Flash~\cite{adobeflash}, and Clip Studio Paint~\cite{clipstudio} to streamline inbetweening and compositing, prominently demonstrated by productions like \textit{Nausicaä of the Valley of the Wind} (1984) and \textit{Neon Genesis Evangelion} (1995). Despite these innovations, digital technologies primarily served as supportive tools rather than automated solutions, still heavily relying on manual artistic inputs.

\paragraph{The AIGC Cel Era (2020s onward).}
Unlike prior technological innovations, Generative AI actively participates in the animation pipeline, significantly automating tasks traditionally requiring intensive human effort. Recent AI-driven methods automate complex processes, including inbetween frame generation, coloring, complex visual effects, and storyboard production~\cite{guajardo2024generativeai,anidoc2024,tooncrafter2024}. For instance, AniDoc~\cite{anidoc2024} employs video diffusion models to automate frame interpolation and colorization, and ToonCrafter~\cite{tooncrafter2024} handles exaggerated non-linear movements and occlusions characteristic of cartoon animations. Platforms like Storyboarder.ai facilitate rapid pre-production by generating detailed storyboards directly from text prompts~\cite{storyboarder}. Practical implementations, such as Netflix Japan’s experimental short film \textit{The Dog and the Boy} (2023), demonstrate the viability of AI-driven workflows in significantly reducing resource requirements. Furthermore, advanced video-generation frameworks like AniSora~\cite{jiang2024anisora} underline the transformative potential of AIGC in enhancing both artistic creativity and production scalability.
\\

Despite significant progress, AIGC integration in animation continues to face critical challenges, including maintaining stylistic consistency, ensuring temporal coherence, and precisely capturing human creative intent~\cite{guajardo2024generativeai,anidoc2024,tooncrafter2024,song2024texttoon}. Addressing these challenges requires systematic examination and thoughtful integration of GenAI tools within established animation practices.

Several recent surveys have explored related areas in generative media production. Li et al.~\cite{li2024survey} discuss challenges in long-video generation, while Lei et al.~\cite{lei2024survey} and Xing et al.~\cite{xing2024survey} examine realistic human motion synthesis and video diffusion models, respectively. Zhou et al.~\cite{zhou2024survey} focus on the broader applications of generative AI in video creation, and Cho et al.~\cite{cho2024survey} extensively review text-to-video generation methods. Zhao et al.~\cite{Zhao2022Cartoon} offer insights into static cartoon image processing with limited exploration of animated contexts.
Zhang et al.~\cite{zhang2025generative} examine the recent advances in GenAI for film creation.
In contrast, our survey explicitly addresses GenAI's integration across the complete Cel-Animation pipeline, examining its impacts on pre-production, production, and post-production phases, alongside critical discussions of existing challenges and promising research directions.

The survey is structured as follows: \Cref{sec:pre} provides foundational knowledge of the traditional Cel-Animation pipeline, covering essential concepts and introducing core GenAI methodologies. \Cref{sec:method} systematically explores GenAI applications across each stage of Cel-Animation production. In \Cref{sec:discussion}, we critically analyze existing challenges and outline potential future directions. Finally, \Cref{sec:conclusion} summarizes our insights and key conclusions.

%% file: sec/2_pre.tex
\section{Preliminary}
\label{sec:pre}

\subsection{Conventional Cel-Animation Pipeline}

\begin{figure}[!ht]
\centering
\includegraphics[width=1\linewidth]{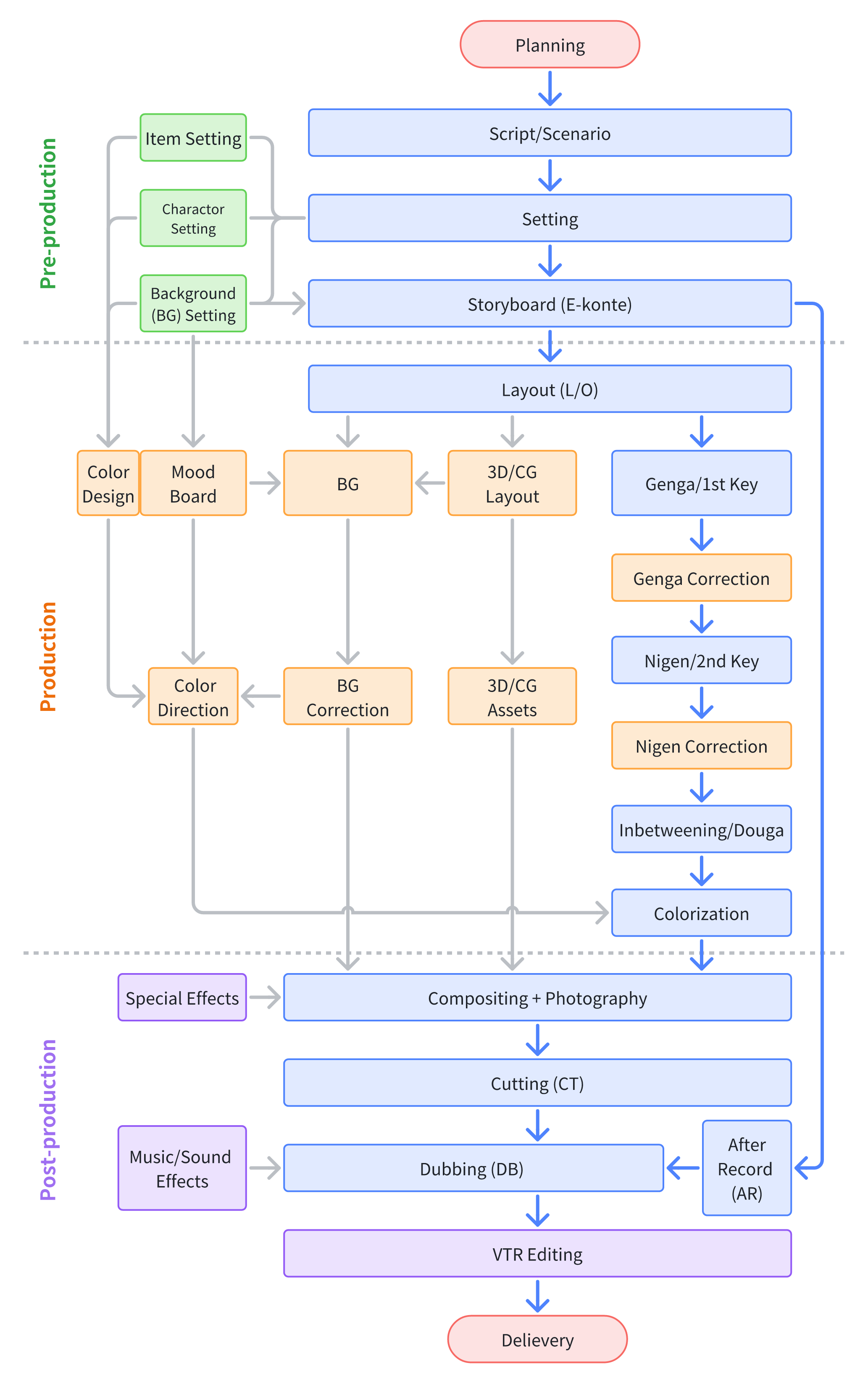}
\caption{Workflow diagram of a traditional animation pipeline.}
\label{fig:pipeline}
\vspace{-1em}
\end{figure}

The Cel-Animation pipeline is a structured process encompassing distinct stages: pre-production, production, post-production, and supplemental processes, each crucial to the animation’s quality and visual appeal (see Figure~\ref{fig:pipeline}).

\subsubsection{\textcolor{preproduction}{\textbf{Pre-production}}}
Pre-production sets the creative and technical foundations for an animation project:

\begin{itemize}
\item \textcolor{preproduction}{\textbf{Scripting}} defines the story, including setting, characters, events, dialogue, and necessary props. Collaboration between directors, writers, and producers is key to establishing narrative direction.
\item \textcolor{preproduction}{\textbf{Setting}} conceptualizes visual elements, including characters, items, and background designs, providing templates for animators to maintain consistency.
\item \textcolor{preproduction}{\textbf{Storyboarding}} visually represents the script through sequential illustrations, determining composition, acting, and timing. Collaboration between directors and storyboard artists finalizes visual storytelling.
\end{itemize}

\subsubsection{\textcolor{production}{\textbf{Production}}}
Production translates pre-production designs into tangible animation assets:

\begin{itemize}
\item \textcolor{production}{\textbf{Layout (L/O)}} specifies framing, perspective, and object relationships based on storyboards, guiding subsequent production phases.
\item \textcolor{production}{\textbf{1st Keyframe Animation (Genga)}} creates rough sketches of key poses and movements, foundational for character animation.
\item \textcolor{production}{\textbf{2nd Keyframe Animation (Nigen)}} refines Genga drawings by adding detailed poses, smoothing transitions, and incorporating corrections from animation directors. Highlight lines are depicted in red; shadow lines in blue.
\item \textcolor{production}{\textbf{Inbetweening}} generates intermediate frames between keyframes, crucial for smooth and continuous motion, often handled by assistant animators.
\item \textcolor{production}{\textbf{Colorization}} applies color based on predetermined models to enhance visual quality and thematic expression, utilizing digital tools for efficiency and precision.
\end{itemize}

\subsubsection{\textcolor{postproduction}{\textbf{Post-production}}}
Post-production consolidates animation components into the final product:

\begin{itemize}
\item \textcolor{postproduction}{\textbf{Compositing \& Photography}}:
Compositing combines characters, backgrounds, and visual effects into finalized scenes. Traditionally involving photography of physical cel layers, modern workflows rely on digital compositing software to achieve similar results.
\item \textcolor{postproduction}{\textbf{Cutting (CT)}} assembles sequences into coherent narratives, aligning scenes for proper timing, pacing, and synchronization.
\item \textcolor{postproduction}{\textbf{Music \& Sound Effects}}:
Sound design includes creating and syncing background music (BGM) and sound effects (SE) to the visuals. These elements enhance emotional depth and provide a more immersive viewing experience.
\item \textcolor{postproduction}{\textbf{After Recording (AR) \& Dubbing (DB)}}:
The dubbing phase involves recording and synchronizing voice acting with the completed animation\footnote{We use the term \textit{dubbing} here to refer to the general pipeline of maintaining consistency between animation and audio/speech, encompassing both AR and traditional dubbing approaches in animation production.}. This step integrates character performances with the visual flow of the animation.
\end{itemize}

\subsubsection{Other Processes}
Additional important processes include:

\begin{itemize}
\item \textbf{Background (BG) Design} creates environments based on settings and layouts, sometimes employing 3D assistance.
\item \textbf{3D Assistance} uses 3D modeling to accurately position elements and virtual cameras, aiding layout accuracy and easing animation complexity.
\item \textbf{Clean-Up} refines preliminary sketches into polished line art, typically performed during 2nd Keyframe Animation.
\item \textbf{Quality Control \& Inspection} includes multiple inspection phases throughout the production to maintain consistent quality, which is also time-consuming.
\end{itemize}

\subsection{Generative AI}

\subsubsection{Large Language Models (LLMs)}
Language models learn the joint probability $p(x_{1:L})$ over a token sequence $x_{1:L}$ via the chain rule:
\begin{equation}
p(x_{1:L})=\prod_{i=1}^{L}p(x_{i}|x_{1:i-1}),
\end{equation}
where $L$ is the sequence length. With billions of parameters, LLMs utilize tokenizers and self-attention layers to predict token probabilities autoregressive:
$\mathcal{M}(x_{1:i-1}) = p(x_i \mid x_{1:i-1})$.
Decoding strategies, such as greedy decoding:
\begin{equation}
x_t = \argmax_{s \in S} \log p_{\mathcal{M}}(s \mid \bm{x}_{1:t-1}),
\end{equation}
control token selection, while sampling strategies promote diversity. Key characteristics of LLMs include:
\begin{itemize}
\item \textit{Scaling Laws}~\cite{kaplan2020scaling}: Model performance grows predictably with increased parameters, data, and compute.
\item \textit{Emergent Abilities}~\cite{zhao2023survey}: At scale, models exhibit novel behaviors like in-context learning, instruction following, and chain-of-thought reasoning.
\end{itemize}
LLMs have been applied in animation and filmmaking to streamline tasks like scriptwriting, character backstory creation, and generating visual descriptions for storyboarding.
MLLMs~\cite{zhu2023minigpt,liu2023visual} extend LLMs by integrating visual encoders and cross-modal aligners, excelling in tasks that require both visual and textual reasoning.

\subsubsection{GAN, VAE, and Diffusion Models}
Generative Adversarial Networks (GANs)~\cite{goodfellow2014generative} consist of generators and discriminators trained adversarially, applicable to background creation and style transfer (StyleGAN~\cite{karras2019style}, Pix2Pix~\cite{isola2017image}).
Variational Autoencoders (VAEs)~\cite{kingma2013auto} optimize latent variable distributions, effective for generating controllable poses and expressions.
Diffusion models~\cite{sohl2015deep,ho2020ddpm} progressively denoise samples via:
\begin{equation}
q(x_t|x_{t-1}) = \mathcal{N}(x_t; \sqrt{1-\beta_t}x_{t-1}, \beta_t \mathbf{I}),
\end{equation}
predicting noise with neural networks $\epsilon_{\theta}(x_t, t)$. Extensions like Stable Diffusion~\cite{rombach2022high} and ControlNet~\cite{zhang2023controlnet} support animation-specific tasks, addressing temporal consistency.

%% file: sec/3_methods.tex
\input{figs/tree}
\section{GenAI for Cel-Animation}
\label{sec:method}

In this section, we discuss how GenAI techniques are tailored to specific processes in Cel-Animation, including those initially designed for other domains but promising for animation workflows.

\subsection{\textcolor{preproduction}{\textbf{GenAI for Pre-production}}}
\label{sec:preproduction}
\paragraph{\textcolor{preproduction}{\textbf{Script Generation.}}}
\label{sec:script-ai}
LLMs facilitate the creation of story settings, plot expansion, and dialogue generation for original animations. For adaptations (e.g., novels, manga, games), LLMs first comprehend source content including narrative structures and character dynamics before adapting or generating scripts. Recent proprietary (ChatGPT~\cite{achiam2023gpt}, Claude~\cite{anthropic_claude35}, Gemini~\cite{reid2024gemini}) and open-source LLMs (LLaMA~\cite{touvron2023llama}, Qwen~\cite{bai2023qwen}) demonstrate substantial adaptability in both original and adapted animation scripts. Frameworks like HoLLMwood~\cite{chen2024hollmwood} automate scriptwriting processes, enhancing narrative coherence and character interaction quality.

\begin{figure}[H]
    \centering
    \includegraphics[width=\linewidth]{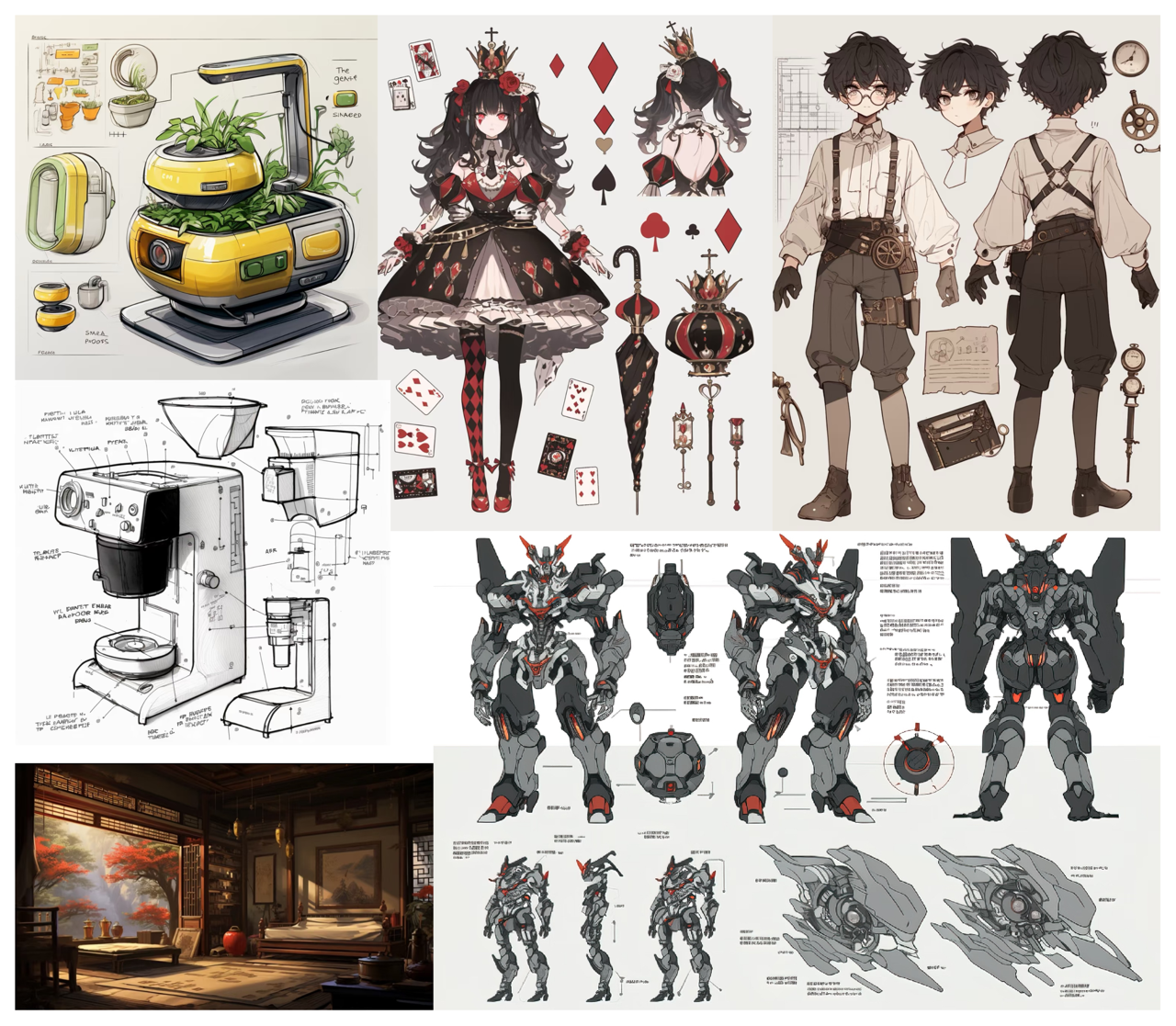}
    \caption{Examples of AI-generated Settings with Midjourney~\cite{midjourney} and Stable Diffusion~\cite{rombach2022high}, showing various designs including items, characters, scenes, and mecha.}
    \label{fig:settei-ai}
\vspace{-1em}
\end{figure}

\paragraph{\textcolor{preproduction}{\textbf{{Setting Generation.}}}}
\label{sec:setting-ai}
Dedicated research for automated setting generation is limited, but practical use cases demonstrate tools such as Midjourney~\cite{midjourney} and Stable Diffusion~\cite{rombach2022high} effectively generating characters, scenes, and objects from text prompts.
As shown in \Cref{fig:settei-ai}, these generated contents demonstrate a certain degree of consistency (for example, multiple views of characters from different angles maintain good identity preservation).
However, detailed inaccuracies remain common, including irrelevant items or nonsensical text within generated designs.

\paragraph{\textcolor{preproduction}{\textbf{Storyboard Generation.}}}
\label{sec:storyboard-ai}
Recent GenAI techniques address automated storyboard generation primarily for live-action films~\cite{lin2023videodirectorgpt} and non-photorealistic animations~\cite{cogcartoon2023, zhang2024storyweaver}. Typically, textual scripts are processed by LLMs into detailed scene descriptions, which then guide image generation models to produce sequential visual narratives. Core challenges include maintaining consistent character identity and ensuring alignment between human intent and AI-generated content. Strategies addressing these challenges involve memory-based identity retention~\cite{rahman2023make, Pan_2024_WACV}, attention-based controls~\cite{yang2024seed}, and semantic alignment enhancements~\cite{shen2024boosting, zhang2024storyweaver}. CogCartoon~\cite{cogcartoon2023} further improves storyboard precision by explicitly controlling character positions within frames.

\subsection{\textcolor{production}{\textbf{GenAI for Production}}}
\label{sec:production}
\paragraph{\textcolor{production}{\textbf{Layout Generation.}}} 
\label{sec:layout-ai}
L/O phase refines rough storyboards by defining character positions, scene composition, perspective, camera angles, and lighting. While some storyboard methods support bounding-box-based spatial control, they often lack fine-grained manipulation of poses and orientations~\cite{li2019layoutgan}. Recent approaches~\cite{zhang2024sketch} improve layout realism by taking sketches as input, enabling control over object size, pose, depth, and perspective. However, these models primarily target object layouts in photorealistic images, limiting effectiveness for anime-specific character layouts. Moreover, layout generation remains largely static. CameraCtrl~\cite{he2024cameractrl} introduces camera motion control in video generation, enabling dynamic layout synthesis with synchronized foreground-background motion and temporal continuity.

\paragraph{\textcolor{production}{\textbf{Keyframe Animation.}}}
\label{sec:keyframe-ai}
Keyframe animation in Cel-Animation focuses on animating scene elements, primarily characters. Recent methods~\cite{hu2023animateanyone, karras2023dreampose} use 2D body landmarks to guide pose generation with spatial fidelity. Systems like Animate-Any-One~\cite{hu2023animateanyone}, Champ~\cite{zhu2024champ}, and MusePose~\cite{musepose} introduce ReferenceNet to bridge source-target pose differences, enhancing transfer accuracy and realism. MimicMotion~\cite{mimicmotion2024} and Animate-X~\cite{tan2024animate} further generalize across body types, including cartoon characters, by modeling morphological variance.
For facial animation, recent works~\cite{song2024texttoon, song2024adaptive, song2021talking, liu2024emo} animate faces using 3D Gaussians or neural representations. TextToon~\cite{song2024texttoon} and Emo-Avatar~\cite{liu2024emo} integrate LLMs with 3D Gaussian splatting for controllable cartoon stylization. Ada-TalkingHead~\cite{song2024adaptive} uses neural keypoints for head animation, while Editable-Head~\cite{song2021talking} adopts landmark-based approaches. These models offer greater control and finer details in facial motion compared to coarse full-body animation methods.


\paragraph{\textcolor{production}{\textbf{Inbetweening.}}}
\label{sec:inbetweening-ai}
While mechanically straightforward, inbetweening is among the most labor-intensive and time-consuming tasks in animation, making it ideal for GenAI acceleration. Accordingly, significant research has emerged in this area. State-of-the-art approaches like ToonCraft~\cite{tooncrafter2024} and AutoFI~\cite{shen2022autofi} adapt video interpolation techniques using diffusion models to generate intermediate frames between keyframes. Other methods incorporate user guidance or structural features: \cite{li2021deep} uses sketches to preserve motion and visual style; SAIN~\cite{shen2024bridging} employs multi-stream U-Transformers to capture region, stroke, and pixel-level cues; AnimeInbet~\cite{siyao2023deep} models line drawings as vector graphs for structure-aware inbetweening. Interactive approaches such as \cite{fukusato2024exploring} provide trajectory-guided sliders for fine control, while \cite{JointStroke} aligns stroke structures across frames. MotionBridge~\cite{tanveer2024motionbridge} further enhances controllability and temporal coherence using structured temporal-spatial representations.


\paragraph{\textcolor{production}{\textbf{Colorization.}}}
\label{sec:colorization-ai}
Colorization is another highly manual process extensively studied in the anime domain. Our focus is on video-level methods using line drawings and sparse color references. Classical models~\cite{anidoc2024} leverage a single colored drawing to guide colorization of black-and-white inputs. Newer methods treat this as a style transfer task, with few-shot approaches like TextToon~\cite{song2024texttoon}, StyleGANEX~\cite{yang2023styleganex}, and VToonify~\cite{yang2022Vtoonify} learning sketch-to-style mappings. However, these often struggle with temporal consistency and generalize poorly beyond avatars. Recent works~\cite{yang2023rerender, yang2024fresco, tokenflow2023} use large pretrained diffusion models to achieve zero-shot, temporally consistent video stylization. Additionally, scene-level control has improved with models~\cite{casey2021animation, huang2024lvcd, Yu2022AnimationColorization, dai2024learning} offering detailed and controllable colorization. Some systems, such as ToonCraft~\cite{tooncrafter2024}, support simultaneous inbetweening and colorization using only sparse input (e.g., one colored frame plus keyframes).

\begin{figure}[!ht]
    \centering
    \includegraphics[width=\linewidth]{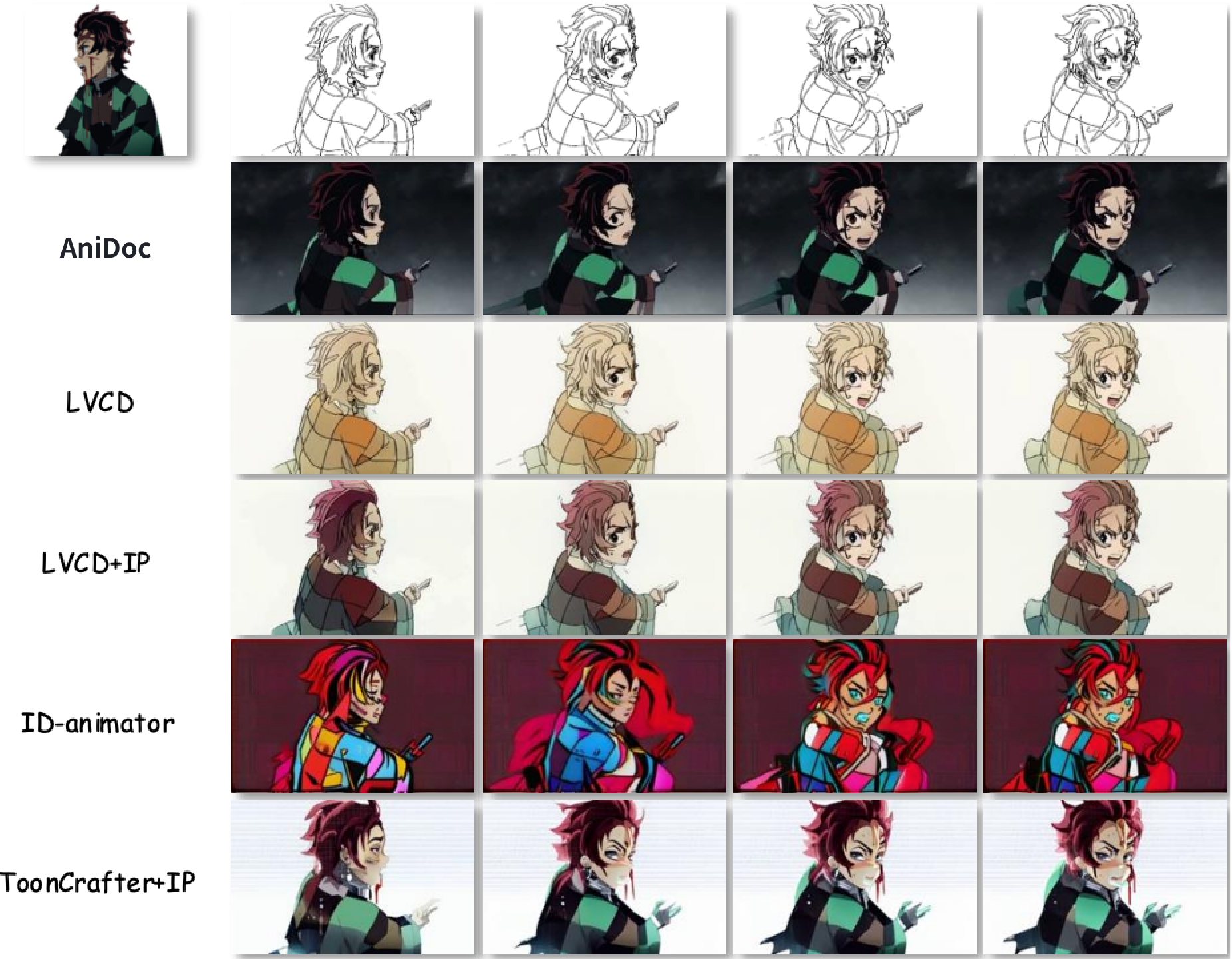}
    \caption{Comparison of colorization generated by GenAI models. The example is adapted from~\cite{meng2024anidoc}.}
    \label{fig:colorization-ai}
\vspace{-1.5em}
\end{figure}

\subsection{\textcolor{postproduction}{\textbf{GenAI for Post-production}}}
\label{sec:postproduction}
\paragraph{\textcolor{postproduction}{\textbf{Compositing \& Photography.}}}
\label{sec:photography-ai}
Compositing requires more than simply overlaying foregrounds on backgrounds; it must ensure consistency in lighting, perspective, and spatial coherence. While layout unifies general geometry, the photography phase handles nuanced refinements, such as matching foreground lighting to scene ambience, adding shadows, and integrating particle or texture effects. Studies~\cite{cong2020dovenet, cong2022high, guerreiro2023pct, jiang2021ssh, tarres2024thinking, man2024floating} explore background-foreground harmonization. For instance, ObjectStitch~\cite{Song2023objectstitch} refines foreground placements for seamless integration, while ICLight~\cite{zhang2024iclight} enables editable, consistent lighting. These techniques collectively advance realism and stylistic coherence in final renders.


\paragraph{\textcolor{postproduction}{\textbf{Music \& Sound Effects.}}}
\label{sec:MusicSoundEffect-ai}
Audio significantly enhances the immersive experience and narrative engagement in animations. Recently, GenAI has made substantial strides in automating audio generation:
\begin{itemize}
    \item \textbf{Music Generation:}
Music in animation often serves to amplify character emotions and enhance the atmosphere. For instance, an energetic song can heighten the intensity of a battlefield scene, while a melancholic tune can evoke empathy and draw the audience into a poignant narrative. AI-driven music generation primarily leverages text-guided methods, including diffusion models~\cite{liu2023audioldm,audioldm2-2024taslp,huang2023noise2music} and sequence-to-sequence modeling~\cite{agostinelli2023musiclm,copet2024simple,huang2024audiogpt}. However, precise synchronization with visual content requires specialized video-guided models such as FoleyMusic~\cite{FoleyMusic2020}, Video2Music~\cite{KANG2024123640}, V2Meow~\cite{su2024v2meow}, MelFusion~\cite{chowdhury2023melfusion}, and VidMuse~\cite{tian2024vidmuse}. Although not specifically tailored for anime, these methods offer critical foundations for synchronization in AI-driven animation.
\item \textbf{Sound Effects Generation:} Effective sound effects require precise alignment with visual cues, encompassing various scene-specific noises (e.g., animal calls, mechanical sounds). Recent transformer-based models (SpecVQGAN~\cite{SpecVQGAN_Iashin_2021}, Im2Wav~\cite{sheffer2022i}, FoleyGen~\cite{mei2024foleygen}) and diffusion-based approaches~\cite{luo2024diff,chen2024action2sound,liang2024language} address semantic and temporal alignment, achieving more realistic and synchronized sound effects.
\end{itemize}



\paragraph{\textcolor{postproduction}{\textbf{Dubbing (DB).}}}
\label{sec:dubbing-ai}
DB synchronizes dialogue audio with character lip movements and expressions, traditionally performed through either pre-animated adjustments or post-animation recording. GenAI approaches, notably EmotiVoice~\cite{cong2024styledubber}, automate this by detecting facial regions and adjusting or generating facial movements guided by reference audio, subtitles, and visual context. Such models fuse acoustic features, phoneme-level precision, and emotional cues to ensure natural and expressive audio-visual synchronization, significantly streamlining the DB process.


\paragraph{\textcolor{postproduction}{\textbf{Cutting.}}}
\label{sec:cutting-ai}


Currently, no dedicated GenAI methods specifically address animation cutting. However, existing general-purpose approaches like Reframe Anything (RAVA)~\cite{cao2024reframe}, which uses LLM-generated FFmpeg instructions for reframing, and M-SAN~\cite{Tang_2022_ACCV}, which extracts coherent segments from longer videos, show potential for adaptation to animation cutting. These methods can aid in efficiently managing duration and preserving content saliency.

\begin{figure*}[!ht]
\centering
\includegraphics[width=1\linewidth]{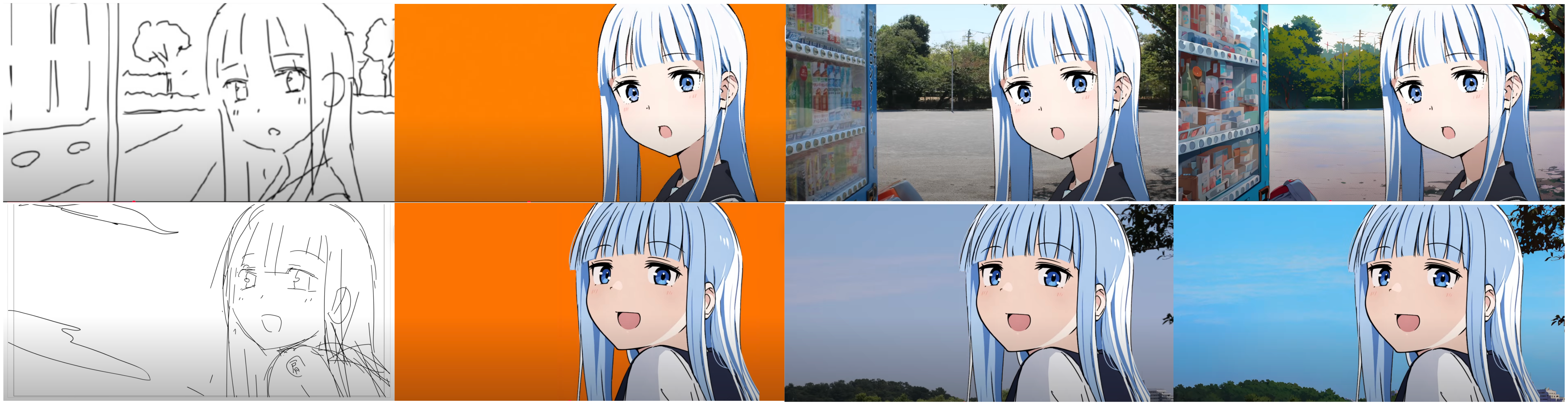}
\caption{Twins HinaHima: Generate cel materials from the storyboard.}
\label{fig:case_hinahima}
\vspace{-1.5em}
\end{figure*}

\subsection{Others}
\label{sec:others-ai}
\paragraph{Cel-Animation Editing.}
\label{sec:editing-ai}
Context-aware editing methods enhance precision and controllability in cel-animation modifications. Re:Draw~\cite{cardoso2024re} blends inpainting and translation methods to ensure local edits remain contextually coherent and visually consistent, utilizing dual discriminators and specialized adversarial losses. Similarly, ScalingConcept~\cite{huang2024scaling}, a training-free editing technique based on pre-trained diffusion models, addresses common post-production issues such as blurred lines and degraded clarity, significantly improving animation visual quality without compromising artistic style.
\paragraph{Cels Decomposition.}
\label{sec:decomposition-ai}
Cel decomposition into reusable sprite layers facilitates effective editing and analysis. Sprite-from-Sprite~\cite{zhang2022sprite} introduces a self-supervised framework handling varying complexity of animated elements. Additionally, Generative Omnimatte~\cite{lee2024generativeOmnimatte} employs generative video priors to robustly handle occlusions, dynamic backgrounds, and environmental effects, thus preserving details like shadows and reflections, further enriching decomposition capabilities.
\paragraph{3D Assistance.}
\label{sec:3d-ai}
AI-assisted 3D methods provide robust tools for enhancing cel-animation dynamics. DrawingSpinUp~\cite{zhou2024drawingspinup} enables automatic 3D animation from character sketches, maintaining artistic integrity while supporting complex movements. Toon3D~\cite{weber2024toon3d} reconstructs cartoon drawings into navigable 3D scenes, allowing novel viewpoints and sophisticated sequence planning. These tools bridge traditional 2D animation with 3D capabilities, offering greater flexibility and richer visual storytelling.


%% file: figs/tree.tex
\tikzstyle{my-box}=[
    rectangle,
    draw=hidden-draw,
    rounded corners,
    text opacity=1,
    minimum height=2em,
    minimum width=5em,
    inner sep=2pt,
    align=center,
    fill opacity=.5,
    line width=0.8pt,
]
\tikzstyle{leaf}=[my-box, minimum height=2em,
    fill=white!80, text=black, align=left,font=\normalsize,
    inner xsep=2pt,
    inner ysep=4pt,
    line width=0.8pt,
]
\begin{figure*}[!ht]
    \centering
    \resizebox{0.85\textwidth}{!}{
        \begin{forest}
            forked edges,
            for tree={
                grow=east,
                reversed=true,
                anchor=base west,
                parent anchor=east,
                child anchor=west,
                base=left,
                font=\large,
                rectangle,
                draw=hidden-draw,
                rounded corners,
                align=left,
                minimum width=6em,
                edge+={darkgray, line width=1pt},
                s sep=20pt,
                inner xsep=3pt,
                inner ysep=3pt,
                line width=0.8pt,
                ver/.style={rotate=90, child anchor=north, parent anchor=south, anchor=center},
            },
            where level=1{text width=7.0em,font=\normalsize,}{},
            where level=2{text width=10em,font=\normalsize,}{},
            where level=3{text width=8.2em,font=\normalsize,}{},
            where level=4{text width=8.2em,font=\normalsize,}{},
            [
                \textbf{GenAI for}\\\textbf{Cel-Animation}\\(\S \ref{sec:method})
                [                
                    \textcolor{preproduction}{\textbf{Pre-production}}\\(\S \ref{sec:preproduction})
                    [
                        \textcolor{preproduction}{\textbf{Scripting}}~(\S \ref{sec:script-ai})
                        [
                            \textcolor{blue_color}{HoLLMwood}~\cite{chen2024hollmwood}{, }
                            \textcolor{blue_color}{ChatGPT}~\cite{achiam2023gpt}{, }
                            \textcolor{blue_color}{Gemini}~\cite{reid2024gemini}{, }
                            \textcolor{blue_color}{Claude}~\cite{anthropic_claude35}{, }\\
                            \textcolor{blue_color}{LLaMA}~\cite{anthropic_claude35}{, }
                            \textcolor{blue_color}{InternVL}~\cite{chen2024internvl}{, }
                            \textcolor{blue_color}{Qwen}~\cite{bai2023qwen}{ }
                            , leaf, text width=26em
                        ]
                    ]
                    [
                        \textcolor{preproduction}{\textbf{Setting}}~(\S \ref{sec:setting-ai})
                        [
                            \textcolor{blue_color}{Midjourney}~\cite{midjourney}{, }
                            \textcolor{blue_color}{Stable Diffusion}~\cite{rombach2022high}{ }
                            , leaf, text width=17em
                        ]
                    ]
                    [
                        \textcolor{preproduction}{\textbf{Storyboarding}}~(\S \ref{sec:storyboard-ai})
                        [
                            {StoryWeaver}~\cite{zhang2024storyweaver}{, }
                            \textcolor{red_color}{CogCartoon}~\cite{cogcartoon2023}{, }
                            \textcolor{red_color}{Coherent Story}~\cite{Pan_2024_WACV}{, }\\
                            \textcolor{red_color}{Seed-Story}~\cite{yang2024seed}{, }
                            \textcolor{red_color}{Make-A-Story}~\cite{rahman2023make}{, }
                            \textcolor{blue_color}{Make-A-Storyboard}~\cite{su2023make}{, }
                            \\                            \textcolor{blue_color}{Storyboarder.ai}~\cite{storyboarder}{, }
                            \textcolor{blue_color}{Animate-A-Story}~\cite{he2023animate-a-story}{, }
                            \textcolor{blue_color}{StoryImager}~\cite{tao2025storyimager}{, }\\
                             \textcolor{blue_color}{VideoDirectorGPT}~\cite{lin2023videodirectorgpt}{, }
                             \textcolor{blue_color}{CustomCrafter}~\cite{wu2024customcrafter}{, }
                             \textcolor{blue_color}{VideoStudio}~\cite{long2024videodrafter}{, }\\
                             \textcolor{blue_color}{Mind the Time}~\cite{wu2024mind}{, }
                             \textcolor{blue_color}{DREAMRUNNER}~\cite{wang2024dreamrunner}{, }
                             \textcolor{blue_color}{Vlogger}~\cite{zhuang2024vlogger}{, }\\
                             {Anim-Director}~\cite{li2024anim}{, }\textcolor{red_color}{StoryGen}~\cite{liu2024storygen}{, }
                            , leaf, text width=28em
                        ]
                    ]
                ]
                [
                    \textcolor{production}{\textbf{Production}}\\(\S \ref{sec:production})
                    [
                        \textcolor{production}{\textbf{Layout (L/O)}}~(\S\ref{sec:layout-ai})
                        [
                            \textcolor{red_color}{CogCartoon}~\cite{cogcartoon2023}{, }
                            \textcolor{blue_color}{SGSIG}~\cite{zhang2024sketch}{, }
                            \textcolor{blue_color}{VideoComposer}~\cite{wang2024videocomposer}{, }\\
                            \textcolor{blue_color}{LayoutGAN}~\cite{li2019layoutgan}{, }
                            \textcolor{blue_color}{Layout2Vid}~\cite{lin2023videodirectorgpt}{, }
                            \textcolor{blue_color}{DiffSensei}~\cite{wu2024diffsensei}{, }\\
                            \textcolor{blue_color}{MangaDiffusion}~\cite{chen2024manga}{, }
                            \textcolor{blue_color}{CameraCtrl}~\cite{he2024cameractrl}{  }
                            , leaf, text width=24em
                        ]
                    ]
                    [
                        \textcolor{production}{\textbf{Keyframe}}\\ \textcolor{production}{\textbf{Animation}}~(\S \ref{sec:keyframe-ai})
                        [
                            AnimateAnyone~\cite{hu2023animateanyone}{, }
                            Champ~\cite{zhu2024champ}{, }
                            MimicMotion~\cite{mimicmotion2024}{, }\\
                            \textcolor{blue_color}{Animate-X}~\cite{tan2024animate}{, }{MikuDance}~\cite{zhang2024mikudance}{, }
                            \textcolor{red_color}
                            {CoNR}~\cite{lin2022collaborative}{  }
                            , leaf, text width=24.5em
                        ]
                    ]
                    [
                        \textcolor{production}{\textbf{Inbetweening}}~(\S \ref{sec:inbetweening-ai})
                        [
                            \textcolor{red_color}{ToonCraft}~\cite{tooncrafter2024}{, }
                            JointStroke~\cite{JointStroke}{, }
                            \textcolor{red_color}{AniSora}~\cite{jiang2024anisora}{, }
                            AnimeInterp~\cite{siyao2021deep}{, }\\
                            AnimeInbet~\cite{fukusato2024exploring}{, }
                            AutoFI~\cite{shen2022autofi}{, }
                            SAIN~\cite{shen2024bridging}{, }
                            MotionBridge~\cite{tanveer2024motionbridge}{, }\\
                            \textcolor{red_color}{toona.io}~\cite{toonaio}{, }
                            DeepSketch-Guided~\cite{li2021deep}{, }
                            DeepGeometrized~\cite{siyao2023deep}{, }\\
                            \textcolor{red_color}{LayerAnimate}~\cite{yang2025layeranimate}{ }
                            , leaf, text width=28.5em
                        ]
                    ]
                    [
                        \textcolor{production}{\textbf{Colorization}}~(\S \ref{sec:colorization-ai})
                        [
                            {AniDoc}~\cite{meng2024anidoc}{, }
                            \textcolor{red_color}{ToonCraft}~\cite{tooncrafter2024}{, }
                            \textcolor{blue_color}{VToonify}~\cite{yang2022Vtoonify}{, }
                            \textcolor{blue_color}{TokenFlow}~\cite{tokenflow2023}{, }\\
                            \textcolor{blue_color}{StyleGANEX}~\cite{yang2023styleganex}{, }
                            \textcolor{blue_color}{FRESCO}~\cite{yang2024fresco}{, }
                            \textcolor{red_color}{toona.io}~\cite{toonaio}{, }TRE-Net~\cite{TRE-Net}{, }\\
                            \textcolor{red_color}{LayerAnimate}~\cite{yang2025layeranimate}{, }{DeepLineArt}~\cite{shi2020deeplineart}{, }{SketchBetween}~\cite{loftsdottir2022sketchbetween}{, }\\
                            Flow Estimate-Refine Net~\cite{Yu2022AnimationColorization}{, }
                            \textcolor{blue_color}{PromptFix}~\cite{yu2024promptfix}{, }MangaNinja~\cite{liu2025manganinja}
                            {, }\\
                            \textcolor{blue_color}{Animation Xformer}~\cite{casey2021animation}{, }
                            Inclusion Matching~\cite{dai2024learning}{, }LVCD~\cite{huang2024lvcd}{ }
                            , leaf, text width=29em
                        ]
                    ]
                ]
                [
                    \textcolor{postproduction}{\textbf{Post-production}}\\(\S \ref{sec:postproduction})
                    [
                        \textcolor{postproduction}{\textbf{Compositing \&}}\\\textcolor{postproduction}{\textbf{Photography}}~(\S \ref{sec:photography-ai})
                        [
                            \textcolor{blue_color}{IC-Light}~\cite{zhang2024iclight}{, }
                            \textcolor{blue_color}{Dovenet}~\cite{cong2020dovenet}{, }
                            \textcolor{blue_color}{High res harmonization}~\cite{cong2022high}{, }\\
                            \textcolor{blue_color}{Pct-net}~\cite{guerreiro2023pct}{, }
                            \textcolor{blue_color}{SSH}~\cite{jiang2021ssh}{, }
                            \textcolor{blue_color}{Object Compositing}~\cite{tarres2024thinking}{, }\textcolor{blue_color}{SSN}~\cite{sheng2021ssn}{, }\\
                            \textcolor{blue_color}{Dr.Bokeh}~\cite{sheng2024dr}{, }
                            \textcolor{blue_color}{Floating No More}~\cite{man2024floating}{, }
                            \textcolor{blue_color}{ObjectDrop}~\cite{winter2024objectdrop}{, }\\
                            \textcolor{blue_color}{Alchemist}~\cite{sharma2024alchemist}{, }
                            \textcolor{blue_color}{Disenstudio}~\cite{chen2024disenstudio}{, }
                            \textcolor{red_color}{LayerAnimate}~\cite{yang2025layeranimate}{ }
                            , leaf, text width=27.5em
                        ]
                    ]
                    [
                        \textcolor{postproduction}{\textbf{Cutting (CT)}}~(\S \ref{sec:cutting-ai})
                        [
                            \textcolor{blue_color}{M-SAN}~\cite{Tang_2022_ACCV}{, }
                            \textcolor{blue_color}{Reframe Anything}~\cite{cao2024reframe}{, }
                            \textcolor{blue_color}{OpusClip}~\cite{opusclip}
                            , leaf, text width=24em
                        ]
                    ]
                    [
                        \textcolor{postproduction}{\textbf{Music \& Sound}}\\\textcolor{postproduction}{\textbf{Effects}}~(\S \ref{sec:MusicSoundEffect-ai})
                        [
                            \textcolor{blue_color}{FoleyMusic}~\cite{FoleyMusic2020}{, }
                            \textcolor{blue_color}{Video2Music}~\cite{KANG2024123640}{, }
                            \textcolor{blue_color}{V2Meow}~\cite{su2024v2meow}{, }\\
                            \textcolor{blue_color}{MelFusion}~\cite{chowdhury2023melfusion}{, }
                            \textcolor{blue_color}{VidMuse}~\cite{tian2024vidmuse}{, }
                            \textcolor{blue_color}{SpecVQGAN}~\cite{SpecVQGAN_Iashin_2021}{, }\\
                            \textcolor{blue_color}{Im2Wav}~\cite{sheffer2022i}{, }
                            \textcolor{blue_color}{FoleyGen}~\cite{mei2024foleygen}{, }
                            \textcolor{blue_color}{DiffFoley}~\cite{luo2024diff}{, }\\\textcolor{blue_color}{Action2Sound}~\cite{chen2024action2sound}{ }
                            , leaf, text width=24em
                        ]
                    ]
                    [
                        \textcolor{postproduction}{\textbf{After Record (AR) \&}}\\ \textcolor{postproduction}{\textbf{Dubbing (DB)}}~(\S \ref{sec:dubbing-ai})
                        [
                            \textcolor{blue_color}{StyleDubber}~\cite{cong2024styledubber}{, }
                            \textcolor{blue_color}{ANIM-400K}~\cite{cai2024anim400k}{,}
                            \textcolor{blue_color}{EmoDubber}~\cite{cong2024emodubberhighqualityemotion}{, }
                            \\
                            \textcolor{blue_color}{MovieDubbing}~\cite{zhang2024from}{, }
                            \textcolor{blue_color}{HPMDubbing}~\cite{cong2023learning}{, }
                            \textcolor{blue_color}{V2C}~\cite{chen2022v2c}{ }
                            , leaf, text width=25em
                        ]
                    ]
                ]
                [
                    Others~(\S \ref{sec:others-ai})
                    [
                        Cel-Animation\\Editing~(\S \ref{sec:editing-ai})
                        [
                            {Re:Draw}~\cite{cardoso2024re}{, }
                            \textcolor{blue_color}{ScalingConcept}~\cite{huang2024scaling}{ }
                            , leaf, text width=17em
                        ]
                    ]
                    [
                        Cels Decomposition\\(\S \ref{sec:decomposition-ai})
                        [
                            Sprite-from-Sprite~\cite{zhang2022sprite}{, }
                            \textcolor{blue_color}{Generative Omnimatte}~\cite{lee2024generativeOmnimatte}{, }\\
                            \textcolor{blue_color}{TransPixar}~\cite{wang2025transpixar}{, }
                            \textcolor{red_color}{LayerAnimate}~\cite{yang2025layeranimate}{ }
                            , leaf, text width=23em
                        ]
                    ]
                    [
                        3D Assistance~(\S \ref{sec:3d-ai})
                        [
                            Toon3D~\cite{weber2024toon3d}{, }
                            \textcolor{red_color}
                            {CoNR}~\cite{lin2022collaborative}{, }{DrawingSpinUp}~\cite{zhou2024drawingspinup}{ }
                            , leaf, text width=23em
                        ]
                    ]
                ]
            ]
        \end{forest}
                                                                           }
    \caption{The taxonomy of GenAI for Cel-Animation is primarily organized by the tasks involved in the production workflow. Key steps are highlighted by \textcolor{preproduction}{Green}, \textcolor{production}{Orange}, and \textcolor{postproduction}{Purple}.
     Methods capable of multiple tasks within the Cel-Animation production are indicated by \textcolor{red_color}{Red}, while \textcolor{blue_color}{Blue} signifies methods that are not originally developed for Cel-related tasks but have the potential to be applied to Cel-Animation.}
    \label{fig:taxonomy}
\vspace{-1.5em}
\end{figure*}

%% file: sec/4_case_study.tex
\section{Case Study}

\paragraph{\textit{The Dog \& The Boy}~\cite{dogboy2022} (2022).}
Netflix Anime Creators Base, WIT Studio, and Rinna introduced a gated latent-diffusion image-to-image pipeline that targets only the Background Design sub-stage.
Layout artists provided line sketches; a model fine-tuned on $\sim$6 k internal BG plates generated color keys that were hand-retouched and composited beneath hand-drawn characters. This focused approach shows how GenAI can shorten the Compositing \& Photography stage while preserving traditional cel animation, but it also revealed open issues regarding data provenance, labor displacement, and authorship in professional discussions.

\paragraph{\textit{Rock, Paper, Scissors}~\cite{corridorRPS2023} (2023).}
Corridor Digital filmed live-action actors on green screen and replaced nearly all assistant work with an AI-assisted rotoscope process. Each frame was styled by a DreamBooth-tuned Stable Diffusion model; a ControlNet edge branch plus optical flow warping ensured temporal consistency, effectively combining Colorization and Inbetweening. The method allowed desktop-scale production but caused flickering and off-model artifacts.

\paragraph{\textit{Twins Hinahima}~\cite{hinahima2025} (2025).} 
Promoted as the first broadcast series with “95 \% AI-assisted cuts,” the show uses a full 3-D to 2-D hybrid. Motion is blocked in Maya/Unreal, rendered in grayscale, then converted to cel frames by multi-ControlNet Stable Diffusion. Per-character LoRA adapters (trained on over 500 images) lock palette and line weight; diffusion-based in-betweening reduces most of the assistant's work. As illustrated in \Cref{fig:case_hinahima}, storyboard sketches can be directly converted into production-quality cels. However, residual face warping and hair-strand drift highlight unresolved challenges in structural fidelity and long-term temporal stability.

%% file: sec/5_discussion.tex
\section{Challenges and Future Directions}
\label{sec:discussion}

In this section, we summarize the main challenges that still prevent reliable studio deployment of GenAI for Cel-Animation and outline directions for future research.

\paragraph{Prompt and Reference Limitations.}
Text prompts capture story beats but miss frame-accurate timing, foreshortened poses, and strict palette discipline. Model sheets, color scripts, and prop libraries prepared in pre-production are usually treated only as loose “style hints,” so faces drift off model and hues wander across cuts. Future systems need to treat every pre-production artifact as a conditioning signal and enforce it as a hard constraint during sampling.
\insightbox{Integrate text, sketches, timing charts, and palette keys into prompts with enforceable constraints.}

\paragraph{Background Leakage and Input Drift.}
Colorization and inbetweening networks often redraw clean-plate backgrounds or distort line geometry provided by key animators, which disrupts multiplane compositing. The problem occurs because latent codes entangle foreground strokes with background textures, so adjusting one unintentionally changes the other. Encoders that separate layers or depth-guided fusion modules can protect clean plates while propagating gradients only through character layers, reducing the need for manual fixes.  
\insightbox{Use layer-aware models so foreground edits do not affect background plates. Objects in different layers should not affect each other.}

\paragraph{Limited Artist Control.}
Prompt-driven generation speeds up idea development but conceals critical manual controls like exposure timing, onion-skin previews, and per-stroke adjustments that professionals need when refining shots. Without clear handles, artists must rerun entire sequences to change a single joint angle or shadow edge. A practical interface would reveal a hierarchy of controls: global style sliders, mid-level tools for pose or camera adjustments, and pixel-perfect masks for final retouching.
\insightbox{Provide hierarchical handles to facilitate predictable and localized edits without necessitating complete regeneration.}

\paragraph{Data Governance and Scale.}
Studio adoption depends on datasets that are both rights-cleared and workflow-complete, meaning storyboards, roughs, clean-ups, color keys, exposure sheets, and final composites are frame-aligned under one license.  
Public corpora only cover fragments and often exist in legal grey areas, which hinders end-to-end training and fair benchmarking. An opt-in consortium that releases full production bundles with standard metadata and provenance tags would provide the coverage and legal clarity needed for progress.
\insightbox{Develop license-compliant, comprehensive datasets to facilitate reproducible training and evaluation.}

%% file: sec/6_conclusion.tex
\section{Conclusion}
\label{sec:conclusion}
This survey examines GenAI's impact on Cel-Animation workflows, tracing its evolution from handcrafted techniques to computer-assisted methods and the current GenAI era. GenAI tools like diffusion models and LMMs enhance efficiency, accessibility, and creativity across all production stages, lowering costs and minimizing repetitive tasks. Tools like AniDoc, ToonCrafter, and AniSora show how automation in storyboard creation, inbetweening, and colorization is changing traditional pipelines, allowing artists to focus more on creativity and innovation.
However, many challenges remain, such as maintaining stylistic consistency, visual coherence, and balancing AI content with human creativity. Ethical concerns and preserving artistic intent are also crucial as the industry adapts. Future GenAI advancements could refine and democratize animation, creating a more inclusive and creative landscape.